\documentclass{sigplanconf}
\usepackage{natbib}
\usepackage{graphicx}
\begin{document}

\setlength{\pdfpageheight}{\paperheight}
\setlength{\pdfpagewidth}{\paperwidth}

\conferenceinfo{CONF 'yy}{Month d--d, 20yy, City, ST, Country}
\copyrightyear{20yy}
\copyrightdata{978-1-nnnn-nnnn-n/yy/mm}
\doi{nnnnnnn.nnnnnnn}

\titlebanner{banner above paper title}        % These are ignored unless
\preprintfooter{short description of paper}   % 'preprint' option specified.

\title{Urdu Handwritten Text Recognition Using ResNet18}

\authorinfo{Muhammad Kashif}
           {National University of Computer and Emerging Sciences}
           {E-mail: I192141@nu.edu.pk}

\maketitle

\begin{abstract}
Handwritten text recognition is an active research area in the field of deep learning and artificial intelligence to convert handwritten text into machine-understandable. A lot of work has been done for other languages, especially for English, but work for the Urdu language is very minimal due to the cursive nature of Urdu characters. The need for Urdu HCR systems is increasing because of the advancement of technology. In this paper, we propose a ResNet18 model for handwritten text recognition using Urdu Nastaliq Handwritten Dataset (UNHD) which contains 3,12000 words written by 500 candidates.
\end{abstract}


\section{Problem statement}
Urdu is the national language of Pakistan and one of the largest languages in the world. Enough amount of research work has been done for other languages but when it comes to Urdu a very limited work has been done especially for handwritten text recognition. The handwritten text has enormous applications for instance converting handwritten files into digital, reading house numbers, postal addresses automatically. National Information and Technology Board (NITB) has introduced E-office in government offices so a system is required to convert all handwritten text into digital format. 
 \begin{figure}[h!]
 \centering
  \includegraphics[width=3in]{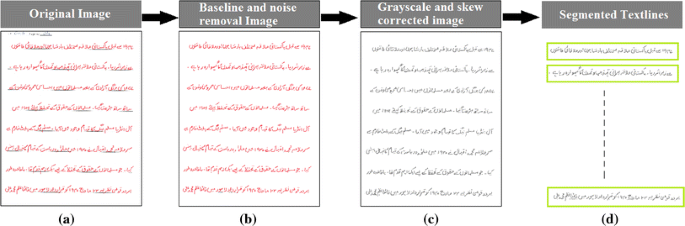}
  \caption{Urdu text converted into machine formate }
\end{figure}

\begin{figure}[h!]
 \centering
  \includegraphics[width=3in]{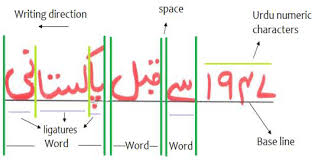}
  \caption{Challenges with urdu text }
\end{figure}

\section{Introduction}
In the field of deep learning and artificial intelligence, handwritten text recognition is one of the challenging tasks because of its cursive nature, and the shape of each character and different writing style is also one of the key issues. 
\begin{figure}[h!]
 \centering
  \includegraphics[width=3in]{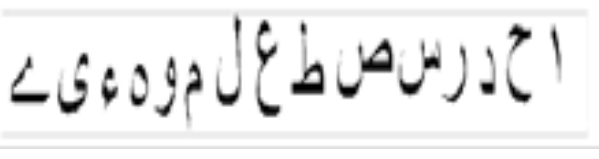}
  \caption{One stroke characters }
\end{figure}
\begin{figure}[h!]
 \centering
  \includegraphics[width=3in]{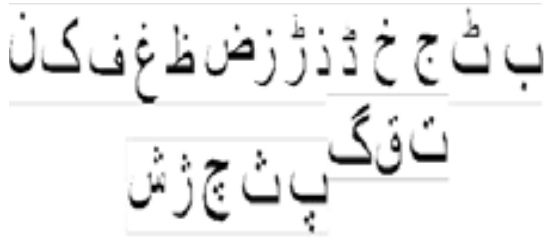}
  \caption{Two and three stroke characters }
\end{figure}
Research work has been done for printed text with higher accuracy rates but there is limited work for handwritten text especially for Urdu handwritten text. Urdu is the official language of Pakistan and in most of the government offices, Urdu is used for writing official orders and directives, documents in government and private organizations in Pakistan. There is a need for converting handwritten text into a machine-understandable because the digital text is safe and we can easily edit it or search it. For example, in handwritten documents/books and digital books, searching a specific topic in digital documents is easy but the same task is not easy in handwritten text. Another example, here in Pakistan we have documents, records of some very sensitive and important data in handwritten files in government offices. One major issue is to store all the files, secondly, if you need some old files, it is very difficult to search for them, and also its time consuming and needs a human resource. But if this data were in digital form it would have been very easy to find old documents and to store them.
\begin{figure}[h!]
 \centering
  \includegraphics[width=3in]{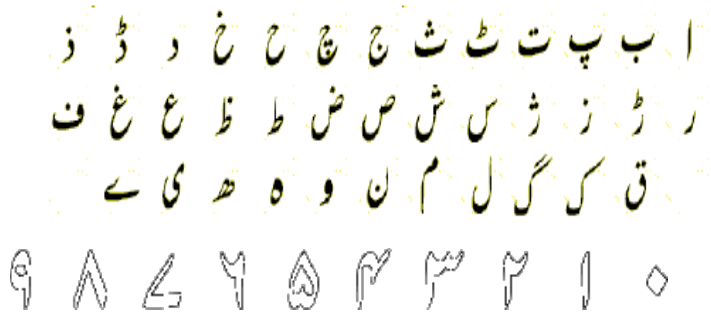}
  \caption{ Urdu Characters and Numerals}
\end{figure}
Motivation: The main reason for this work is that Urdu is an official language of Pakistan and one of the largest languages of the world. secondly, very limited research has been carried out for handwritten text in Urdu. Thirdly, converting handwritten text into digital has enormous applications such as reading postal addresses, reading house numbers, and reading signboards for driverless cars, also a huge amount of data in government offices are in a handwritten format, saving or keeping handwritten text is difficult as it needs a separate place and someone to maintain all these documents and there is a high chance of losing this data in disasters but keeping these documents in digital format, one can save a lot of time and resource. The hand-written text is not like a typical text in one single font, but recognition of handwritten text is a challenging task due to its writing styles vary from person to person. \\ Background: Optical Character Recognition (OCR) is a technology that picks out handwritten or printed text from digital images of handwritten text or printed text for example scanned documents. Optical Character Recognition examines the digital images and generates a code for the text that can be used to retrieve the text from the image into digital format. Optical Character Recognition also referred to as text recognition that is used to convert physical documents into machine-readable text. OCR has typically four different types i.e. Input acquisition mode (online or offline text), written mode (handwritten or printed), character connectivity (isolated or joined), and font constraints (single or Omni font). These types of Optical Character Recognition are shown in figure 6 \cite{khan2018urdu}.
\begin{figure}[h!]
 \centering
  \includegraphics[width=3in]{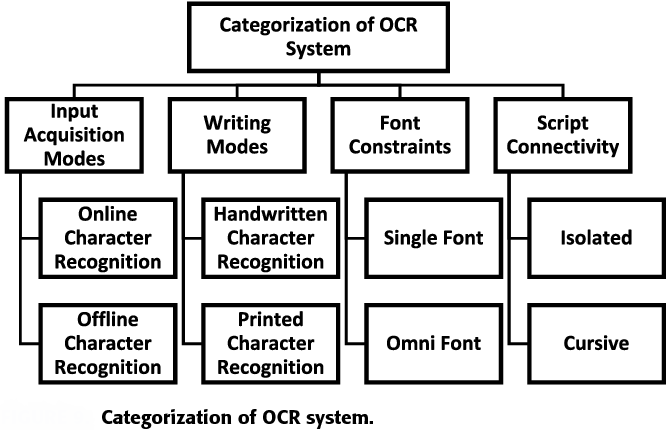}
  \caption{OCR System. }
\end{figure}
\begin{enumerate}
\item {\bf Input acquisition Mode} 
Input acquisition mode is the mode in which an input is given to the optical character recognition. It is of two types i.e. Online recognition and Offline recognition.
\item {\bf Writing Mode} 
The writing mode resources available for OCR can be in any format that is typically written text which contains tables, headers, footers, borders, etc. and another format can be a handwritten format which has different styles of writing for different people. 
\item {\bf Font Constraints}
Font constraints are the type that is developed for one font style. It may fail or partially succeed for text written in another style. Optical Character Recognition that is only for a single font style is known as font constraints OCR.  
\item {\bf character connectivity} 
Another type of Optical Character Recognition is character connectivity which is isolated or joined characters. isolated characters are those that are not joined with each other, while joined characters are joined and it changes the shape and position of a character.
\end{enumerate}
A handwritten text system is a system that recognizes a handwritten text pattern and converts the user's handwritten characters or words into a format that the computer understands. It has many applications in to-days era i.e. reading home addresses and reading numbers automatically or reading traffic signs for driverless cars. Without this technology, we have dependent on handwriting texts that result in errors. Handwritten text or data needs to update manually and it is difficult to save, maintain or search for a specific file. It also needs someone to update it manually. Storing handwritten data is difficult, people have lost important data because of disasters, and they no backup of this handwritten data.\\ Urdu is the national language of Pakistan and one of the largest languages in the world. it is widely used in government offices, institutions, banks, and very important documents are in handwritten form and it has many applications in this technology era such as reading home numbers, reading billboards, etc.  Limited research has been done for Urdu text recognition mostly for printed text. For handwritten text, a research is being done but there is no system available to provide a reliable result.\\ UNHD Database, finding a data set for Urdu Nastaliq handwriting is a separate challenge. There are very few free available datasets that are large enough to train a model. In this paper, we use Urdu Nastaliq Handwritten Dataset (UNHD), which has Urdu numbers, digits, etc, but in Urdu Nasta’liq font. It covers commonly used ligatures that are written by 500 writers with their handwriting on A4 size paper. The dataset has 3,12000 words written by 500 people with a total of 10,000 lines \cite{ahmed2019handwritten}.
\begin{figure}[h!]
 \centering
  \includegraphics[width=3in]{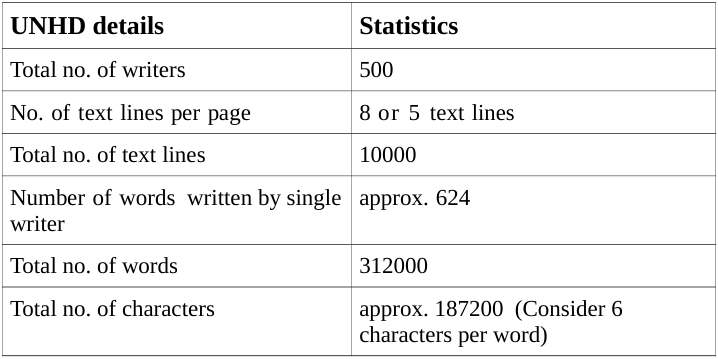}
  \caption{UNHD dataset details}
\end{figure}
\section{Related work}
Related work for Urdu handwritten text recognition is limited as compared to other languages, but people are working for the Urdu language, and it's a good research area. Urdu text style is usually connected and has many variations therefore it is a challenging task. The literature shows that for character-level recognition, the artificial neural network (ANN) is widely used. AN ANN is based on the concept of working of a human brain, it is a collection of nodes called artificial neurons \cite{van2017artificial}. These nodes can send input from one layer to another. Designing an OCR that can recognize multiple languages is challenging as every language exhibits different characteristics and features. Generalizing this kind of issue is nearly impossible. To address this problem a new approach was proposed by Ali et al. \cite{ali2004language} shows a new technique for the character of a language and presented all the characters of a language as a geometrical stroke. Their technique is font independent but it can also apply to standard fonts. The specialty of their technique is that their model needs to be trained once. Figure 8 shows the geometrical strokes of Urdu characters. The accuracy rate is 70\% to 80\% with 25 samples of handwritten text. This technique has some limitations that it will not work on words having dots and signs.  
\begin{figure}[h!]
 \centering
  \includegraphics[width=3in]{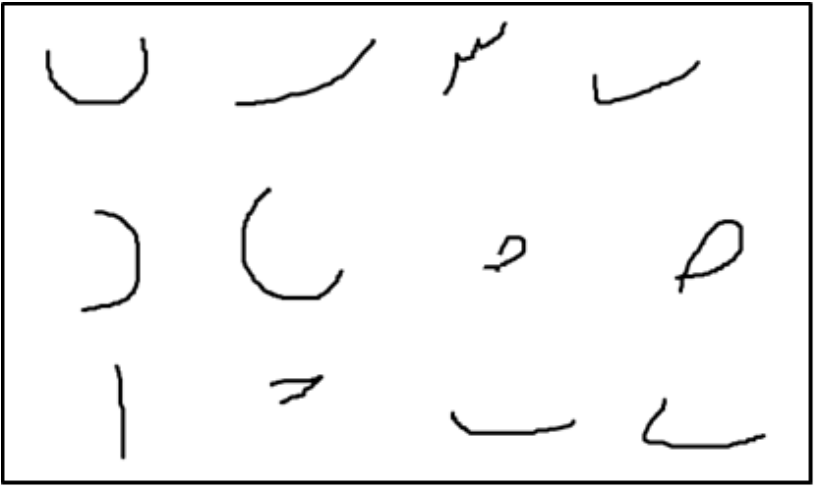}
  \caption{Basic geometrical strokes of Urdu script}
\end{figure}
The Urdu language has a large character set, it has the issue of similarity of strokes. To address this issue, Haider et al. \cite{khan2010online} proposed a methodology for dividing a set of Urdu characters into groups as shown in figure 9.
\begin{figure}[h!]
 \centering
  \includegraphics[width=3in]{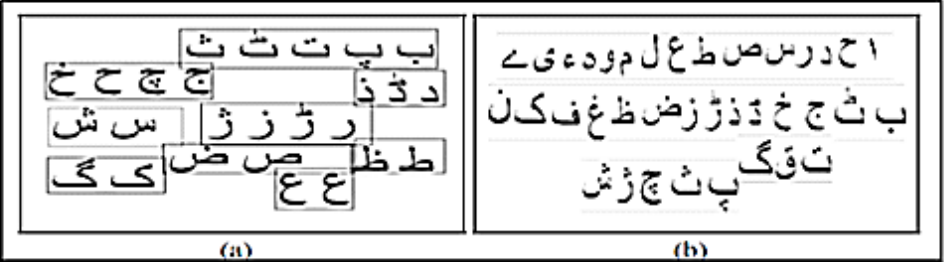}
  \caption{Urdu characters (a) is matching order (b) is set of groups of urdu characters.}
\end{figure}
For single-character recognition, Shahzad et al. \cite{shahzad2009urdu} presented a technique that set up features when using isolated characters by examining the keystrokes and secondary strokes. A linear classifier was used for a total of 190 characters which was written by different writers also contains 38 Urdu characters. 6\% error was detected by the classifier due to the similarity of characters.
\begin{figure}[h!]
 \centering
  \includegraphics[width=3in]{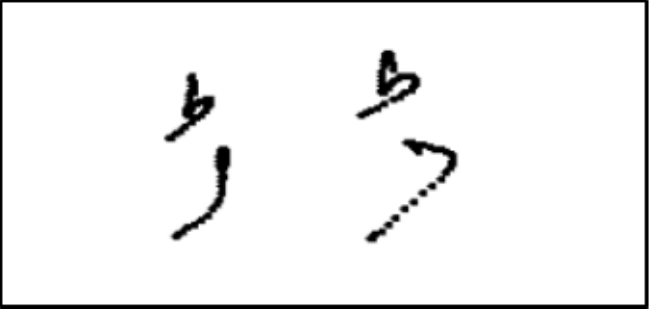}
  \caption{Similar characters "De" (left) and "Daal" (right)}
\end{figure}
Similar work has been proposed by Khan et al. \cite{khan2014online} considering characters that change their shape and position. A multidimensional neural network is widely used in a multidimensional environment for sequence learning and decomposition \cite{husnain2019recognition}. The Urdu Printed Text Image (UPTI) has been used for assessment purposes, which has 10,000 images of handwritten and printed text \cite{sabbour2013segmentation}.\\ In \cite{razzak2009multi} \cite{razzak2009numeral}, the authors presented a similarity between Urdu and Arabic language handwritten data. They used the logic of HMM for Urdu and Arabic to recognize numerals. 97\%, 96\%, and 97.8\% accuracies are obtained by using the fuzzy logic rule.\\ Recent work: Ahmad et al.\cite{ahmad2017offline} has developed a system for Offline Urdu Nastaleeq Based on Stacked Denoising Autoencoder. Similarly, Rizvi et al. \cite{rizvi2019optical} has proposed a technique for Nastalique like Urdu text by using supervised learning. Their system has obtained the highest rate ever achieved by Optical character recognition that is 98.4\% training and 97.3\% test results by many experimental settings. A comprehensive Urdu dataset named Urdu Nastaliq Handwritten Dataset (UNHD) is presented by Ahmed et al.\cite{ahmed2019handwritten}. The dataset contains commonly used ligatures and is written by 500 people on A4 size paper. This is one of the largest datasets of Urdu which are available free and can be accessed by emailing them. Chhajro et al.\cite{chhajro2020handwritten} uses different techniques of deep learning to recognize Urdu handwritten characters through pictures. The different algorithm has been used for evaluation via pictures such as Multilayer perceptron (MLP), K-Nearest neighbor, Recurrent neural network, etc. Ali et al.\cite{ali2020pioneer} has introduced a dataset for Urdu handwritten characters which is written by 900 people and they have used a deep autoencoder for recognition of these characters. Husnain et al.\cite{husnain2020urdu} provides a survey of different techniques for Urdu text recognition, which shows interesting facts. Figure 11 shows details of Urdu datasets. 
\begin{figure}[h!]
 \centering
  \includegraphics[width=3in]{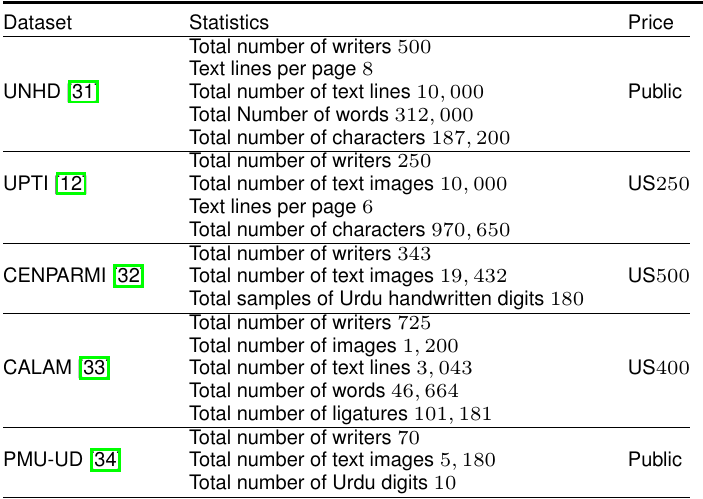}
  \caption{Details of Urdu datasets}
\end{figure}
\begin{figure}[h!]
 \centering
  \includegraphics[width=3in]{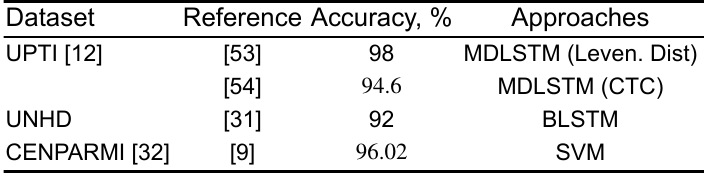}
  \caption{rates reported on common datasets}
\end{figure}
\begin{figure}[h!]
 \centering
  \includegraphics[width=3in]{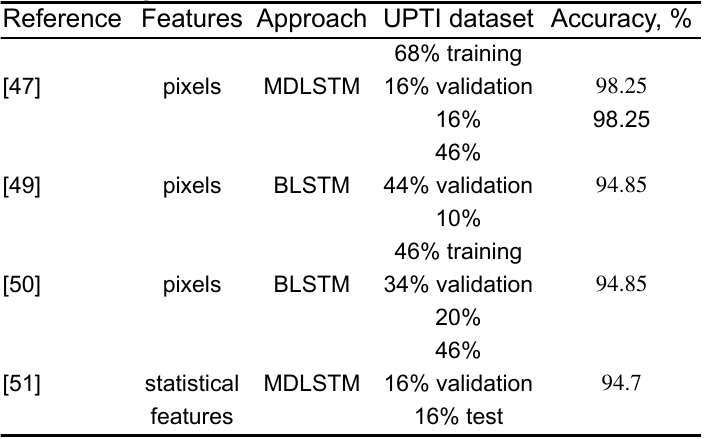}
  \caption{comparison results on UPTI dataset}
\end{figure}
Figure 14 shows year-wise published articles distribution.
\begin{figure}[h!]
 \centering
  \includegraphics[width=3in]{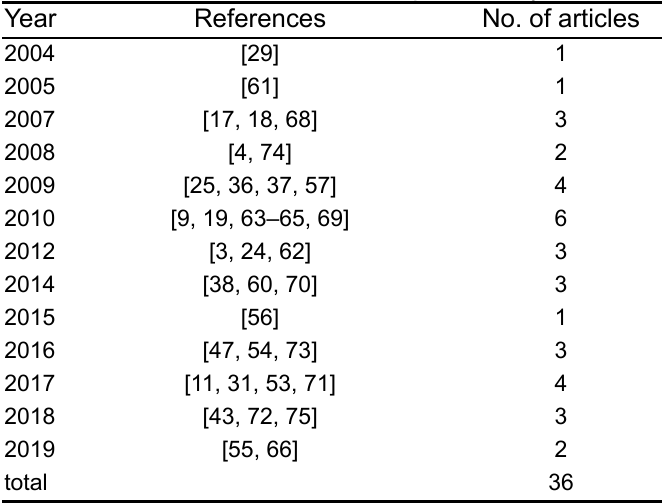}
  \caption{Year wise distribution of related papers}
\end{figure}

In \cite{husnain2019recognition}, proposed a technique for recognition of multi-font Urdu handwritten characters written by different people from different age groups and institutions. figure 15 shows the different isolated Urdu characters.

\begin{figure}[h!]
 \centering
  \includegraphics[width=3in]{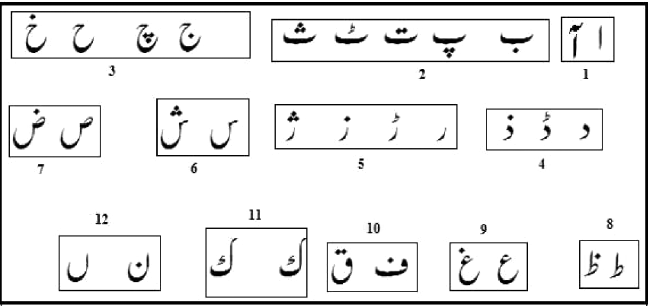}
  \caption{Urdu characters grouped by their shape}
\end{figure}
In \cite{zahid2020roman}, they have proposed a dataset named RU dataset from social media data. This data is important for companies and institutions for future prediction. The dataset is manually interpreted where each sentence contains sentiment labels. \cite{majeed2020emotion} has worked for emotion detection in roman Urdu as these days people express their emotions on social media by posting and sharing pictures and statuses. Some work has been done for emotion detection but in roman Urdu, limited work has been done. They have applied different algorithms such as SVM, KNN, random forest. \cite{naeem2020deep} has presented a technique for detecting fake news, yellow journalism, and social media as these days it is very easy to post or share fake news and there is a huge amount of this kind of contents so to manage it intelligently there is a need for a model that detects these kinds of news and sensationalism. They have obtained an accuracy of 97\% by using Long Short-Term Memory (LSTM). \cite{arshad2019corpus} has presented an evaluation technique for detecting emotion in a short text. Emotion detection has many applications i.e. intelligent agents, smart clinical diagnoses, etc. They have collected 10,000 sentences in roman Urdu to experiment with roman Urdu. \cite{qamarrelationship} has presented a methodology for relationship identification using emotion on social media in the form of text messages or posts. The use of social media has become a part of our day-to-day life, people express their emotions in the form of posts, statuses, or text messages. Using these data they have proposed a method of Relationship Identification using Emotion Analysis  (RIEA). They have obtained 85\% accuracy using RIEA. \cite{javed2020collaborative} has proposed a framework for health and fitness evaluation of people by using machine learning techniques. The framework is called Collaborative Shared Healthcare Plan (CSHCP). The results show a satisfactory outcome. \cite{beg2019algorithmic} has applied machine learning techniques to predict trading decisions using data of stock markets which generate a huge amount of data every day. In \cite{farooq2019melta} they have presented a method level energy estimation MELTA for android applications to compute the energy consumed by android applications. \cite{sahar2019towards} has proposed a methodology for computation of energy consumption by structural information related to energy consumption. As Energy is one of the main concerns of the developers these days. They have proposed three object-oriented suits. \cite{farooq2019bigdata} has used a machine learning technique for energy-related posts on stack overflow. They have analyzed 4 million posts from stack overflow using the machine learning technique NLTK. In \cite{zafar2019constructive}, they have presented a restriction-based generator for the generation of that work for general games. Their results are quite good and can be compared with other level generators. In \cite{dilawar2018understanding}, the authors presented a machine learning technique to obtain a suitable representation of the social media data to help governments with people opinion and their problems related to local issues, political, religion. This will help the government to know about people's opinions and their problems. They have obtained the highest F1 score of 76.40\% in SemEval 2016 task 5 and for SemEval 2015 task 12 they have achieved 94.99\%. In \cite{awan2021top}, They have proposed a top rank approach to extract key phrases from documents. The approach extracts the position of key phrases from the document and enlarges it with topical key phrases. The top rank attains an F1 score of 0.73 for key phrases which is ahead of other state-of-the-art approaches. In \cite{javed2020alphalogger}, They have developed an approach Alpha logger that deduces the alphabet keys typed on a keypad. Typing on a keypad of smartphones generates vibrations that can be used to recognize the keys. In \cite{alvi2017ensights}, They have presented a tool named Ensight that helps software developers by giving them energy consumption details. The tool has been tested on three open-source android applications and gives the F1 score of 86\%.        
\section{Methodology}
\begin{itemize}
\item	Our initial approach was to apply 1-Dimensional Bidirectional LSTM on the UNHD dataset, we got the dataset but couldn't get the desired results. We then applied a ResNet18 Model on Urdu Nastaliq Handwritten Dataset (UNHD). the Resnet model is a residual network used for computer vision tasks. ResNet is based on skip connection. The figure below shows how Resnet uses skip connection. 
\begin{figure}[h!]
 \centering
  \includegraphics[width=3in]{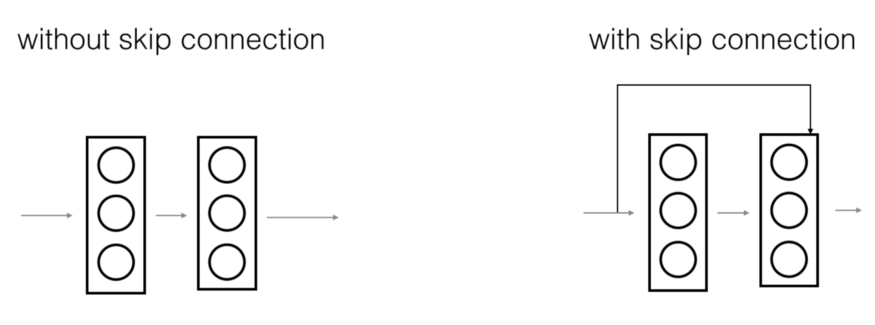}
  \caption{Skip Connection }
\end{figure}

\begin{figure}[h!]
 \centering
  \includegraphics[width=3in]{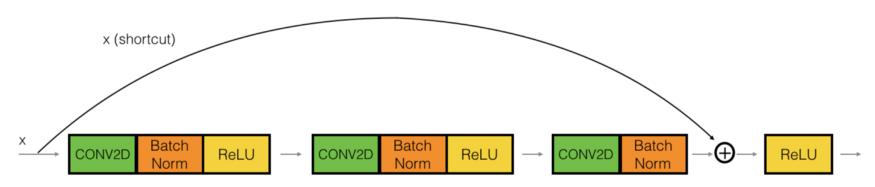}
  \caption{Block diagram }
\end{figure}

\begin{figure}[h!]
 \centering
  \includegraphics[width=3in]{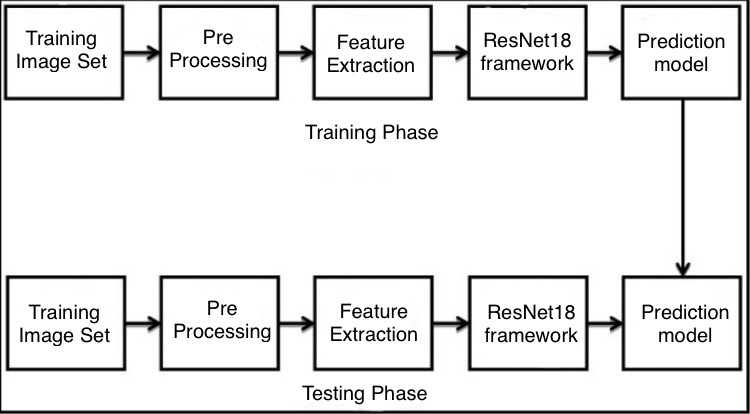}
  \caption{Block diagram of proposed ResNet }
\end{figure}
\begin{itemize}
\item {\bf Pre Processing}: 
Pre-processing is the stage when images go through pre-processing for further processing. The image was first processed to remove noise. 
\item {\bf Feature Extraction}: 
In this stage, the Urdu handwritten characters are processed to extract features such as the shape of the character, the height of the character, and horizontal and vertical lines of the image.
\item {\bf ResNet18 Framework}: Resnet18 applies 18 convolution layers on dataset images to learn and predict the output.
\end{itemize}
Our contribution to this work is that we applied the ResNet18 framework which has 18 convolution layers on Urdu Nastaliq Handwritten Dataset (UNHD), to the best of our knowledge we applied ResNet18 on UNHD for the first time. So a new framework applied to the Urdu dataset, its a good learning work for new students in this field and us as well.  
\item	The idea was to recognize Urdu handwritten text with better accuracy than previous work. First to work on any of the previously proposed techniques and comes up with a better result if it does not work, then apply a new technique on a different dataset. 
\item	The proposed model is significant because It has 18 convolution layers so it recognizes Urdu text more in-depth because of its shape and cursive style.
\end{itemize}
\section{Evaluation and Experiments}
\begin{itemize}
\item	The experiment is conducted on a Lenovo laptop with an Intel® Core™ i7-10750H CPU @ 2.60GHz, 16 Gigabyte of ram, with a graphic card NVIDIA GEFORCE GTX 1660 Ti, Window 10 operating system. Colab was used for coding with Python as a programming language. The UNHD dataset was split as a training set and test set.  
\item	We consider a proposed work already done on the UNHD dataset with the highest accuracy rate as our benchmark because the goal was to come up with more accurate results than previous works.
\item This benchmark is appropriate for our results because the dataset matters a lot in recognition of handwritten text, so we consider a work that is done on the same dataset.
\end{itemize}

\acks
I would like to thank Almighty Allah who gave me the strength and courage to accomplish this project in this trying time.\\ I am very thankful to my advisor Dr. Mirza Omer Beg for his supervision, help, and his amazing lectures throughout the course. \\I am also thankful to all my friends for help. 

\bibliographystyle{plain}
\bibliography{Reference.bib}

\begin{thebibliography}{10}

\bibitem{ahmad2017offline}
Ibrar Ahmad, Xiaojie Wang, Ruifan Li, and Shahid Rasheed.
\newblock Offline urdu nastaleeq optical character recognition based on stacked
  denoising autoencoder.
\newblock {\em China Communications}, 14(1):146--157, 2017.

\bibitem{ahmed2019handwritten}
Saad~Bin Ahmed, Saeeda Naz, Salahuddin Swati, and Muhammad~Imran Razzak.
\newblock Handwritten urdu character recognition using one-dimensional blstm
  classifier.
\newblock {\em Neural Computing and Applications}, 31(4):1143--1151, 2019.

\bibitem{ali2004language}
Anjum Ali, Mohmood Ahmad, Nasir Rafiq, Javed Akber, Usman Ahmad, and Shahwar
  Akmal.
\newblock Language independent optical character recognition for hand written
  text.
\newblock In {\em 8th International Multitopic Conference, 2004. Proceedings of
  INMIC 2004.}, pages 79--84. IEEE, 2004.

\bibitem{ali2020pioneer}
Hazrat Ali, Ahsan Ullah, Talha Iqbal, and Shahid Khattak.
\newblock Pioneer dataset and automatic recognition of urdu handwritten
  characters using a deep autoencoder and convolutional neural network.
\newblock {\em SN Applied Sciences}, 2(2):1--12, 2020.

\bibitem{alvi2017ensights}
Hamza~M Alvi, Hareem Sahar, Abdul~A Bangash, and Mirza~O Beg.
\newblock Ensights: A tool for energy aware software development.
\newblock In {\em 2017 13th International Conference on Emerging Technologies
  (ICET)}, pages 1--6. IEEE, 2017.

\bibitem{arshad2019corpus}
Muhammad~Umair Arshad, Muhammad~Farrukh Bashir, Adil Majeed, Waseem Shahzad,
  and Mirza~Omer Beg.
\newblock Corpus for emotion detection on roman urdu.
\newblock In {\em 2019 22nd International Multitopic Conference (INMIC)}, pages
  1--6. IEEE, 2019.

\bibitem{awan2021top}
Mubashar~Nazar Awan and Mirza~Omer Beg.
\newblock Top-rank: a topicalpostionrank for extraction and classification of
  keyphrases in text.
\newblock {\em Computer Speech \& Language}, 65:101116, 2021.

\bibitem{beg2019algorithmic}
Mirza~O Beg, Mubashar~Nazar Awan, and Syed~Shahzaib Ali.
\newblock Algorithmic machine learning for prediction of stock prices.
\newblock In {\em FinTech as a Disruptive Technology for Financial
  Institutions}, pages 142--169. IGI Global, 2019.

\bibitem{chhajro2020handwritten}
MA~Chhajro, H~Khan, F~Khan, K~Kumar, AA~Wagan, and S~Solangi.
\newblock Handwritten urdu character recognition via images using different
  machine learning and deep learning techniques.
\newblock {\em Indian Journal of Science and Technology}, 13(17):1746--1754,
  2020.

\bibitem{dilawar2018understanding}
Noman Dilawar, Hammad Majeed, Mirza~Omer Beg, Naveed Ejaz, Khan Muhammad, Irfan
  Mehmood, and Yunyoung Nam.
\newblock Understanding citizen issues through reviews: A step towards data
  informed planning in smart cities.
\newblock {\em Applied Sciences}, 8(9):1589, 2018.

\bibitem{farooq2019bigdata}
Muhammad~Umer Farooq, Mirza~Omer Beg, et~al.
\newblock Bigdata analysis of stack overflow for energy consumption of android
  framework.
\newblock In {\em 2019 International Conference on Innovative Computing
  (ICIC)}, pages 1--9. IEEE, 2019.

\bibitem{farooq2019melta}
Muhammad~Umer Farooq, Saif Ur~Rehman Khan, and Mirza~Omer Beg.
\newblock Melta: A method level energy estimation technique for android
  development.
\newblock In {\em 2019 International Conference on Innovative Computing
  (ICIC)}, pages 1--10. IEEE, 2019.

\bibitem{husnain2020urdu}
Mujtaba Husnain, Malik Muhammad~Saad Missen, Shahzad Mumtaz, Micka{\"e}l
  Coustaty, Muzzamil Luqman, and Jean-Marc Ogier.
\newblock Urdu handwritten text recognition: a survey.
\newblock {\em IET Image Processing}, 14(11):2291--2300, 2020.

\bibitem{husnain2019recognition}
Mujtaba Husnain, Malik~Muhammad Saad~Missen, Shahzad Mumtaz, Muhammad~Zeeshan
  Jhanidr, Micka{\"e}l Coustaty, Muhammad Muzzamil~Luqman, Jean-Marc Ogier, and
  Gyu Sang~Choi.
\newblock Recognition of urdu handwritten characters using convolutional neural
  network.
\newblock {\em Applied Sciences}, 9(13):2758, 2019.

\bibitem{javed2020alphalogger}
Abdul~Rehman Javed, Mirza~Omer Beg, Muhammad Asim, Thar Baker, and Ali~Hilal
  Al-Bayatti.
\newblock Alphalogger: Detecting motion-based side-channel attack using
  smartphone keystrokes.
\newblock {\em Journal of Ambient Intelligence and Humanized Computing}, pages
  1--14, 2020.

\bibitem{javed2020collaborative}
Abdul~Rehman Javed, Muhammad~Usman Sarwar, Mirza~Omer Beg, Muhammad Asim, Thar
  Baker, and Hissam Tawfik.
\newblock A collaborative healthcare framework for shared healthcare plan with
  ambient intelligence.
\newblock {\em Human-centric Computing and Information Sciences}, 10(1):1--21,
  2020.

\bibitem{khan2014online}
Kamran~Ullah Khan et~al.
\newblock Online urdu handwritten character recognition: Initial half form
  single stroke characters.
\newblock In {\em 2014 12th International Conference on Frontiers of
  Information Technology}, pages 292--297. IEEE, 2014.

\bibitem{khan2010online}
Kamran~Ullah Khan and Ihtesham Haider.
\newblock Online recognition of multi-stroke handwritten urdu characters.
\newblock In {\em 2010 International Conference on Image Analysis and Signal
  Processing}, pages 284--290. IEEE, 2010.

\bibitem{khan2018urdu}
Naila~Habib Khan and Awais Adnan.
\newblock Urdu optical character recognition systems: Present contributions and
  future directions.
\newblock {\em IEEE Access}, 6:46019--46046, 2018.

\bibitem{majeed2020emotion}
Adil Majeed, Hasan Mujtaba, and Mirza~Omer Beg.
\newblock Emotion detection in roman urdu text using machine learning.
\newblock In {\em Proceedings of the 35th IEEE/ACM International Conference on
  Automated Software Engineering Workshops}, pages 125--130, 2020.

\bibitem{naeem2020deep}
Bilal Naeem, Aymen Khan, Mirza~Omer Beg, and Hasan Mujtaba.
\newblock A deep learning framework for clickbait detection on social area
  network using natural language cues.
\newblock {\em Journal of Computational Social Science}, pages 1--13, 2020.

\bibitem{qamarrelationship}
Saira Qamar, Hasan Mujtaba, Hammad Majeed, and Mirza~Omer Beg.
\newblock Relationship identification between conversational agents using
  emotion analysis.
\newblock {\em Cognitive Computation}, pages 1--15.

\bibitem{razzak2009multi}
Muhammad~Imran Razzak, SA~Hussain, Abdel Bela{\"\i}d, and Muhammad Sher.
\newblock Multi-font numerals recognition for urdu script based languages.
\newblock {\em International Journal of Recent Trends in Engineering (IJRTE)},
  2009.

\bibitem{razzak2009numeral}
Muhammad~Imran Razzak, SA~Hussain, and Muhammad Sher.
\newblock Numeral recognition for urdu script in unconstrained environment.
\newblock In {\em 2009 International Conference on Emerging Technologies},
  pages 44--47. IEEE, 2009.

\bibitem{rizvi2019optical}
SSR Rizvi, A~Sagheer, K~Adnan, and A~Muhammad.
\newblock Optical character recognition system for nastalique urdu-like script
  languages using supervised learning.
\newblock {\em International Journal of Pattern Recognition and Artificial
  Intelligence}, 33(10):1953004, 2019.

\bibitem{sabbour2013segmentation}
Nazly Sabbour and Faisal Shafait.
\newblock A segmentation-free approach to arabic and urdu ocr.
\newblock In {\em Document recognition and retrieval XX}, volume 8658, page
  86580N. International Society for Optics and Photonics, 2013.

\bibitem{sahar2019towards}
Hareem Sahar, Abdul~A Bangash, and Mirza~O Beg.
\newblock Towards energy aware object-oriented development of android
  applications.
\newblock {\em Sustainable Computing: Informatics and Systems}, 21:28--46,
  2019.

\bibitem{shahzad2009urdu}
Nabeel Shahzad, Brandon Paulson, and Tracy Hammond.
\newblock Urdu qaeda: recognition system for isolated urdu characters.
\newblock In {\em Proceedings of the IUI Workshop on Sketch Recognition,
  Sanibel Island, Florida}. Citeseer, 2009.

\bibitem{van2017artificial}
Marcel Van~Gerven and Sander Bohte.
\newblock Artificial neural networks as models of neural information
  processing.
\newblock {\em Frontiers in Computational Neuroscience}, 11:114, 2017.

\bibitem{zafar2019constructive}
Adeel Zafar, Hasan Mujtaba, Sohrab Ashiq, and Mirza~Omer Beg.
\newblock A constructive approach for general video game level generation.
\newblock In {\em 2019 11th Computer Science and Electronic Engineering
  (CEEC)}, pages 102--107. IEEE, 2019.

\bibitem{zahid2020roman}
Rabail Zahid, Muhammad~Owais Idrees, Hasan Mujtaba, and Mirza~Omer Beg.
\newblock Roman urdu reviews dataset for aspect based opinion mining.
\newblock In {\em 2020 35th IEEE/ACM International Conference on Automated
  Software Engineering Workshops (ASEW)}, pages 138--143. IEEE, 2020.

\end{thebibliography}

\end{document}